%% file: root.tex
\documentclass[letterpaper, 10 pt, journal, twoside]{Resources/ieeetran}

\title{\titled}

\author{John N. Nganga$^{1}$ and Patrick M. Wensing$^{1}$\\[-9.5ex]

\thanks{$^{1}$Authors are with the Department of Mechanical Engineering, University of Notre Dame,
        Notre Dame, IN 46556, USA
        {\tt\small \{jnganga,pwensing\}@nd.edu}}%
\thanks{Digital Object Identifier (DOI): see top of this page.}

}

\usepackage{titlesec}

\usepackage{lipsum}
\usepackage{verbatim} 
\usepackage{pgffor}
\usepackage{cite}
\usepackage{Resources/commands}
\usepackage{Resources/qting}
\usepackage{xcolor}

\begin{document}

\maketitle
~
\vspace{-5ex}

\markboth{SEE OFFICIAL VERSION AT \url{http://dx.doi.org/10.1109/LRA.2021.3098928}}
{Nganga and Wensing: Accelerating Second-Order DDP For Rigid-Body Systems} 

\begin{abstract}
This letter presents a method to reduce the computational demands of including second-order dynamics sensitivity information into the Differential Dynamic Programming (DDP) trajectory optimization algorithm. An approach to DDP is developed where all the necessary derivatives are computed with the same complexity as in the iterative Linear Quadratic Regulator~(iLQR). Compared to linearized models used in iLQR, DDP more accurately represents the dynamics locally, but it is not often used since the second-order derivatives of the dynamics are tensorial and expensive to compute. This work shows how to avoid the need for computing the derivative tensor by instead leveraging reverse-mode accumulation of derivative information to compute a key vector-tensor product directly. \quotelabel{rev:abstract}{\rev{We also show how the structure of the dynamics can be used to further accelerate these computations in rigid-body systems.}} Benchmarks of this approach for trajectory optimization with multi-link manipulators show that the benefits of DDP can often be included without sacrificing evaluation time, and can be done in fewer iterations than iLQR.
\end{abstract}
\begin{IEEEkeywords}
Optimization and Optimal Control; Underactuated Robots; Whole-Body Motion Planning and Control
\end{IEEEkeywords}

\section{Introduction}
\IEEEPARstart{I}{n} recent years, online optimal control strategies have gained widespread interest in many applications from motion planning of robots to control of chemical processes~\cite{wieber2016modeling,diehl2009efficient,mayne2000constrained, MultiShooting,Collocation,centroidalDyn}. 
Rather than relying on manually derived policies, these control strategies optimize a metric of cost that encodes desired task goals. This approach then allows online control performance that is generalizable across tasks or environments. 
\quotelabel{rev: BadExample}{\rev{For example, online optimization may enable legged systems to tailor their gaits to sensed terrains and inevitable disturbances or may enable manipulators to rapidly synthesize efficient motions when transporting new objects.}}

However, for robots with even a few links, the underlying system dynamics are complex, nonlinear, and expensive to evaluate. These features challenge the ability to solve trajectory optimization problems online,
particularly in systems with a large number of degrees of freedom~(DoFs). 
Yet, the motivation to perform online optimization is often greater for these very systems, since a high DoF morphology gives the needed flexibility and mobility to adapt to a wider range of situations. Since the curse of dimensionality precludes the ability of exploring the full state space, online optimal control strategies often settle on exploring within a local neighborhood. 
Even then, the optimization of trajectories is often orders of magnitude slower than real-time. For many years, control approaches in the legged robotics literature have sidestepped this burden by employing simple models to enable faster computation \cite{wieber2016modeling}.

\begin{figure}[tb]
    \centering
    \vspace{-2ex}
    \includegraphics[width = .9\columnwidth]{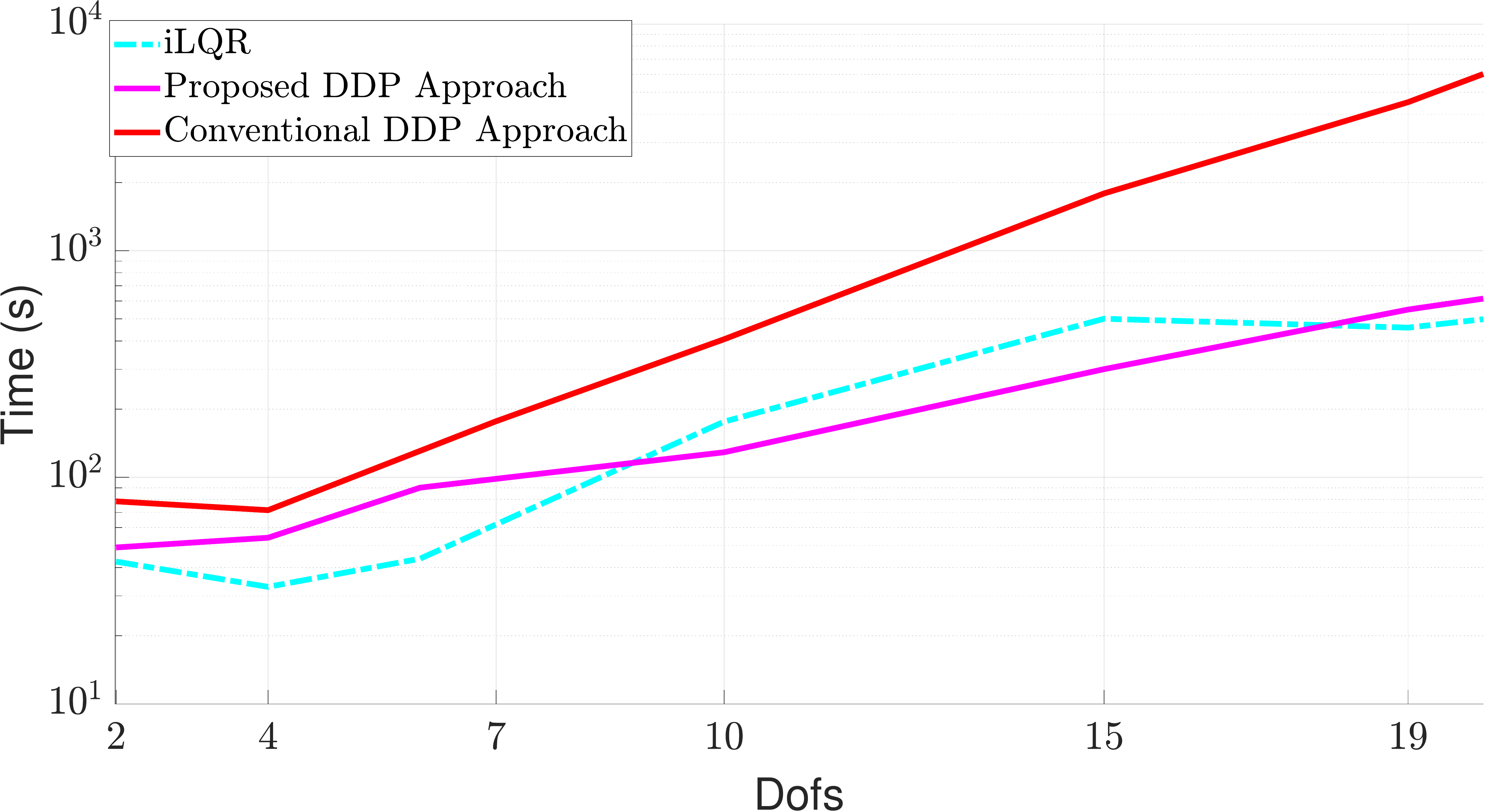}
    \caption{Trajectory optimization time in {\sc Matlab} for an underactuated serial manipulator swing-up task vs.~DoFs. \rev{All cases start from an initial guess of a dissipative controller.} Pink curve: Proposed tensor-free second-order DDP. Red curve: Second-order DDP with explicit computation of second-order dynamics derivative tensors. Blue curve: iLQR method that approximates DDP using first-order dynamics derivatives to save computation time.}
    \label{fig:Fig1}
    \vspace{-2.5ex}
\end{figure}

Recently,  whole-body trajectory optimization is becoming more feasible and has gained increased interest as computation power progresses \cite{koenemann2015whole,neunert2018whole,grandia2019feedback,MultiShooting}.  For example, \cite{tassa2012synthesis} used DDP~\cite{mayne1966DDP} for a humanoid to perform complex tasks such as getting up from an arbitrary pose. DDP 
exploits the sparsity of an optimal control problem (OCP), and its output includes an optimal trajectory along with a locally optimal feedback policy that can be used to handle disturbances \cite{grandia2019feedback}. 
While DDP  natively does not address constraints, many recent approaches using Augmented Lagrangian \cite{lantoine2012hybrid,howell2019altro}, interior point \cite{InteriorPointDDP}, and relaxed barrier strategies \cite{grandia2019feedback,li2020hybrid} have been proposed to handle general state and control constraints, with specialized approaches considered for control limit constraints \cite{tassa2014control,sola2020squash}. 
Other work~\cite{pellegrini2017multiple,giftthaler2018family,plancher2018performance,farshidian2017real} has considered multi-threading and parallelization of the DDP algorithm to accelerate its computation. Finally, Li et al.~\cite{li2020model} combine the advantages of whole-body DDP and simple models by sequentially considering both over the horizon.
Collectively, these previous works show broad potential impact from advances to numerical methods for DDP.


Originally described in~\cite{mayne1966DDP}, DDP uses a second-order approximation of the dynamics when constructing a second-order approximation of the optimal cost-to-go.
However, in practice~(e.g., \cite{tassa2012synthesis, li2004iterative,koenemann2015whole}) many researchers have opted to use a first-order dynamics approximation due to its faster evaluation time, giving rise to the iterative Linear Quadratic Regulator (iLQR). While the second-order dynamics information retains higher fidelity to the full model locally, it is represented by a rank three tensor and is expensive to compute. In this work, we alleviate these computational demands by describing a new approach that avoids the evaluation of the  dynamics derivative tensor (\figref{fig:Fig1}).

    
\subsection{Specific Contributions}\label{sect: Contributions}
\quotelabel{rev: Contributions}{\rev{ 
This work presents and combines several advances for reducing the computational complexity of computing second-order dynamics sensitivity information in DDP. The final result is a method for computing this information with the same computational complexity as first-order dynamics derivatives  (e.g., as in iLQR). The contributions are (I) the use of reverse-mode automatic differentiation~(AD) \cite{grievank2000principles} to compute second-order derivatives needed in DDP. This contribution is general to discrete-time dynamic systems and enables computation reductions compared to methods that explicitly evaluate a derivative tensor for the dynamics. 
Further, we show (II) how second-order information related to the forward dynamics can be related to associated information from the inverse dynamics, akin to first-order results in \cite{carpentier2018analytical,jain1993linearization}; this contribution is specific to rigid-body models. Lastly, we (III) introduce a modification to the Recursive-Newton-Euler Algorithm (RNEA) that supports this process and further reduces computational demands.
%
Figure~\ref{fig:Fig1} overviews a benchmark of the proposed methods against iLQR and against DDP approaches that explicitly compute derivative tensors. }}

\section{Trajectory Optimization via DDP/iLQR}\label{sect: DDPBackground}
This work considers the efficient solution of a finite-horizon OCP for a rigid-body system such as an articulated robot. This section reviews background on dynamics and trajectory optimization with a focus on the DDP algorithm.
\subsection{Dynamics}
The inverse dynamics~(ID) of a rigid-body system are 
\begin{equation}
\t = \M(\q)\qdd + \C(\q,\qd) + \tg(\q) \triangleq \InvDyn(\q,\qd,\qdd,\a_g)\,, 
\label{eq:id}
\end{equation}
where $\M(\q)$ represents the mass matrix, $\C(\q,\qd)$ \rev{Coriolis and centrifugal terms}, and $\tg(\q)$ the gravity term with $\a_g$ the gravitational acceleration. 
We group the Coriolis, centrifugal, and gravity terms together as $\h(\q,\qd) = \C(\q,\qd) + \tg(\q)$. 
The RNEA \cite{originalRNEA1979,featherstone2014rigid} 
can evaluate $\InvDyn$ 
with \Ord{n} complexity where $n$ is the number of DoFs in the system. 
The forward dynamics~(FD) of the system can be formulated as
\[
 \qdd = \M(\q)^{-1} \left(\t - \h(\q,\qd)\right) \triangleq \FwdDyn(\q,\qd,\t, \a_g)\,.
\]
When the fourth argument is omitted for $\InvDyn$ or $\FwdDyn$, gravity of $9.81$ m/s$^2$ downward is assumed.
The Articulated-Body Algorithm~(ABA)~\cite{featherstone2014rigid} can compute $\FwdDyn$ in \Ord{n} complexity and is an efficient alternative to \Ord{n^3} algorithms that calculate and invert the mass matrix to carry out $\FwdDyn$~(e.g.,~\cite{OriginalMassMatrix}).
Continuous trajectories for the state $\x = [\q^T,\qd^T]^T$ and control input $\t$ are discretized herein using a numerical integration scheme. While the strategies proposed are applicable for use with any explicit integration scheme, forward Euler integration is assumed to simplify the remaining development such that:  
\begin{align}\label{eqn:DiscreteDyn}
\begin{split}
\x_{k+1} &= \f(\x_k,\u_k) \triangleq \x_k + h \begin{bmatrix} \qd \\ \FwdDyn(\q,\qd,\t) \end{bmatrix}  
\end{split}\,,
\end{align}
where $h$ is the integration stepsize. 

\subsection{\rev{Differential Dynamic Programming}}\label{sect:DDP}

Herein, DDP \cite{mayne1966DDP} and iLQR \cite{li2004iterative} are used to solve an OCP with a cost function of the form
\begin{equation}\label{eqn: cost-to-go}
    V_0(\x_0,\U_0)=\l_f (\x_N) + \sum_{k=0}^{N-1} \l_k(\x_k, \u_k)
\end{equation}
where $\l_k( \x_k,\u_k)$ represents the running cost, 
 $\l_f( \x_N)$ represents the terminal cost incurred at the end of a 
horizon, and  $\U_0=[\u_0,\u_1,\ldots,\u_{N-1}]$ is the control sequence over the horizon.
A cost-to-go function $V_k(\x_k, \U_k)$ can be similarly defined from any time point as the partial sum of costs from time $k$ to $N$. The cost-to-go function $V_0( \x_0, \U_0)$ in \eqref{eqn: cost-to-go} is minimized with respect to $\U_0$, with states subject to the discrete system dynamics~\eqref{eqn:DiscreteDyn}, providing
\[
  \U_0^\star( \x_0) = \argmin{\U_0} V_0( \x_0, \U_0).
\]
Throughout the paper, the star superscript refers to an optimal value. 
DDP optimizes over each control $\u_i$ separately, working backwards in time by approximating Bellman's equation
\begin{align} \label{eqn:ValueFnWithQ}
\begin{split}
 V^\star_k( \x_k) &= \min_{\u_k} \left[ \underbrace{\l_k(\x_k,\u_k) + V^{\star}_{k+1} (\f( \x_k,\u_k)) }_{Q_k( \x_k,\u_k)} \right] \\
 & {\textrm{ where}} \quad V^\star_N(
 \x_N) = \l_f (\x_N)\,.
 \end{split}
\end{align}
Consider a perturbation to 
$Q_k$ due to small perturbations around a nominal state-control pair $\bar{ \x}_k$ and $\bar{\u}_k$ such that 
\[
\delta Q_k(\delta  \x_k,\delta \u_k)=Q_k(\bar{\x}_k+\delta \x_k, \bar{\u}_k + \delta \u_k) - Q_k(\bar{\x}_k, \bar{\u}_k) \,.
\]
Expanding the $\delta Q$-function to second order leads to
\begin{equation}{\label{eqn:Q_fn}}
\delta Q_k
\approx \frac{1}{2} 
{\begin{bmatrix}
1\\\delta \x_k\\\delta \u_k
\end{bmatrix}^{T}}
{\begin{bmatrix}
0 & \Qx^{T} & \Qu^{T}\\\Qx & \Qxx & \Qux^T \\\Qu & \Qux & \Quu
\end{bmatrix}}
{\begin{bmatrix}
1\\\delta \x_k\\\delta \u_k
\end{bmatrix}}
\end{equation} 
where subscripts indicate partial derivatives. Omitting the time index $k$ for conciseness, these quantities are given by:
\begin{subequations}\label{eqn:coeffs}
\begin{align}
    \Qx &= \l_\x + \fx^{T}\Vx' \tag{\theequation a} \\
    \Qu &= \l_\u + \fu^{T}\Vx' \tag{\theequation b} \\
    \Qxx &= \lxx + \fx^{T}\Vxx'\fx + \Vx'\cdot \fxx \tag{\theequation c} \label{QxxCoef}\\
    \Quu&=\luu + \fu^{T} \Vxx '\fu + \Vx'\cdot \fuu\tag{\theequation d} \label{QuuCoef}\\
    \Qux &=\lux+\fu^{T}\Vxx'\fx + \Vx'\cdot \fux\;. \tag{\theequation e} \label{QuxCoef}
\end{align}
\end{subequations} 
The prime in~\eqref{eqn:coeffs} denotes the next time step, i.e.\textbf{}, $\Vxx' = \Vxx(k+1)$. The last terms in~(\ref{QxxCoef} - \ref{QuxCoef})
denote contraction with a tensor and are ignored in iLQR, representing the main difference between DDP and iLQR. \quotelabel{rev: HamClarify}{\rev{These coefficients~\eqref{eqn:coeffs} could alternatively be viewed through derivatives of the Hamiltonian
\begin{equation}
H_k(\x,\u,\llambda) =  \l_k(\x, \u) + \llambda^T \f(\x,\u)\,,
\label{eq:hamiltonian}
\end{equation} 
where $\llambda \triangleq \Vx'$ is the co-state vector. For example,\cite{mayne1966DDP}
\begin{align*}
\Qx  &= \nabla_{\x} H_k (\x_k, \u_k,\Vx')\quad {\textrm{ and}} \\ 
\Qux &= \nabla^2_{\u\x} H_k (\x_k, \u_k,\Vx') + \fu^{T}\Vxx'\fx \, .
\end{align*}}}
Minimizing~\eqref{eqn:Q_fn} over $\delta \u_k$ attains the incremental control
\begin{align}\label{eqn: optimalControl}
\begin{split}
    \delta \u^{\star}_k &=\argmin{\delta \u_k} \delta \Qk (\delta  \x_k,\delta \u_k) \\
    &=-\underbrace{ \Quu^{-1} \Qu}_{:=\kkappa_k} -\underbrace{\Quu^{-1} \Qux}_{:=\K_k}\delta \x_k\,,
\end{split}
\end{align}
where $\delta \x = \x - \bar{\x}$ denotes the deviation from the nominal state. 
When this control is substituted in \eqref{eqn:Q_fn}, the quadratic approximation of the value function \rev{can be constructed as}
\begin{subequations}\label{eqn:quadValue}
\begin{align}
    \textrm{ER}(k) &= \frac{1}{2}\Qu^T\Quu^{-1}\Qu + \textrm{ER}(k+1)\tag{\theequation a}\\
    \Vx(k) &= \Qx - \Qu^T\Quu^{-1}\Qux\tag{\theequation b}\\
    \Vxx(k) &=\Qxx - \Qxu \Quu^{-1}\Qux, \tag{\theequation c}
\end{align}
\end{subequations}
where the $\textrm{ER}(k)$ is the expected reduction in cost-to-go if $\delta \x_k = 0$ and $\U_k$ were chosen optimally. This process is repeated until a value function approximation is obtained at time $k=0$, constituting the backward sweep of DDP.  

Following this backward sweep, a forward sweep proceeds by simulating the system forward in time under the incremental control policy \eqref{eqn: optimalControl}, resulting in a new state-control trajectory \cite{mayne1966DDP}. This optimal control law~\eqref{eqn: optimalControl} is typically modified by a backtracking line-search parameter, $0 < \epsilon < 1$ such that $\u_k = \bar{\u}_k - \epsilon \kkappa_k - \K_k \delta \x_k$. The line-search parameter ensures that DDP/iLQR takes steps that result in a reduction of the total cost. The resulting trajectory serves as a new nominal trajectory, with the above backward and forward sweeps repeated until some convergence criteria is met. 
\subsection{DDP and iLQR: Conceptual Comparison}

Many factors influence the relative performance of iLQR and DDP, with the main difference again being that the tensorial terms in~(\ref{QxxCoef} - \ref{QuxCoef}) are ignored in iLQR. 
The effect is that while DDP experiences quadratic convergence for trajectories that are sufficiently close to local optimality \cite{tassa2012synthesis}, iLQR only experiences super-linear convergence (i.e., it converges more slowly).
However, if the running and terminal costs are strictly convex, iLQR can be simplified relative to DDP since the  terms ~(\ref{QxxCoef} - \ref{QuxCoef}) in iLQR then ensure that $\Quu$ is always positive definite. By comparison, the addition of the tensor terms~(\ref{QxxCoef} - \ref{QuxCoef}) in DDP can render $\Quu$ indefinite, requiring  regularization~\cite{liao1991convergence,tassa2012synthesis}, which incurs additional computational cost. 
The choice of DDP and iLQR in application then becomes a cost-benefit analysis among these differences with iLQR favored in recent work~\cite{tassa2012synthesis,li2004iterative,koenemann2015whole}. The derivatives of the dynamics are the most computationally expensive terms in DDP, motivating methods for their efficient evaluation. 

\section{\rev{Efficient Computation of Second-Order Derivatives for DDP}}\label{sect:Derivs}

\quotelabel{rev:costate}{\rev{This section considers how dynamics partials enter the partials of the Hamiltonian \eqref{eq:hamiltonian}. The co-state vector is partitioned as $\llambda = [ \xxi\T, \eeta\T ]^T$ where $\xxi$ and $\eeta$ are the co-states associated with $\q$ and $\qd$ respectively. Via \eqref{eqn:DiscreteDyn}, the Hamiltonian is then:
\begin{equation}
    \!\! H_k(\x,\u,\llambda) = \l_k(\x,\u) + \Vector\! \left( \x + \insideArg  \right). \nonumber
\end{equation}}}
Focusing on the second-order partials of $H_k(\x,\u,\llambda)$ with respect to $\q$, and dropping the arguments for conciseness, \quotelabel{rev:partialofqd}{\rev{the partials can be written as
\begin{align}
\pdd[H]{  q_i}{  q_j} &= \pdd[\l]{q_i}{  q_j} + h \Vector \pd[]{q_i} \left(  \pd[]{q_j} \insideArgDD \right) \nonumber\\
&= \pdd[\l]{q_i}{  q_j} + h \eeta\T \pd[]{q_i}\left[ \pd[]{q_j}   \FwdDyn \right]\,,  \label{eqn: Hqq}
\end{align}
where the simplification occurs since $\partial \qd / \partial q_j=0$.}}

In~(\ref{eqn: Hqq}), the second-order partial of $\FwdDyn$ is the most expensive term to compute. It represents a bottleneck in DDP since computing it for all possible $i$ and $j$ results in a tensor $\frac{\partial^2 \FwdDyn}{\partial \q^2}$ with $n^3$ elements. These elements can be computed with total complexity $\Ord{n^3}$ before being contracted with $\eeta$ at an additional $\Ord{n^3}$ total cost. This conventional strategy is denoted as {\emph \TensorCont}. 


\subsection{Reverse-mode Accumulation to Efficiently Compute \eqref{eqn: Hqq}}\label{sect:GeneralAD}
\begin{figure}
    \centering
    \includegraphics[width = \columnwidth]{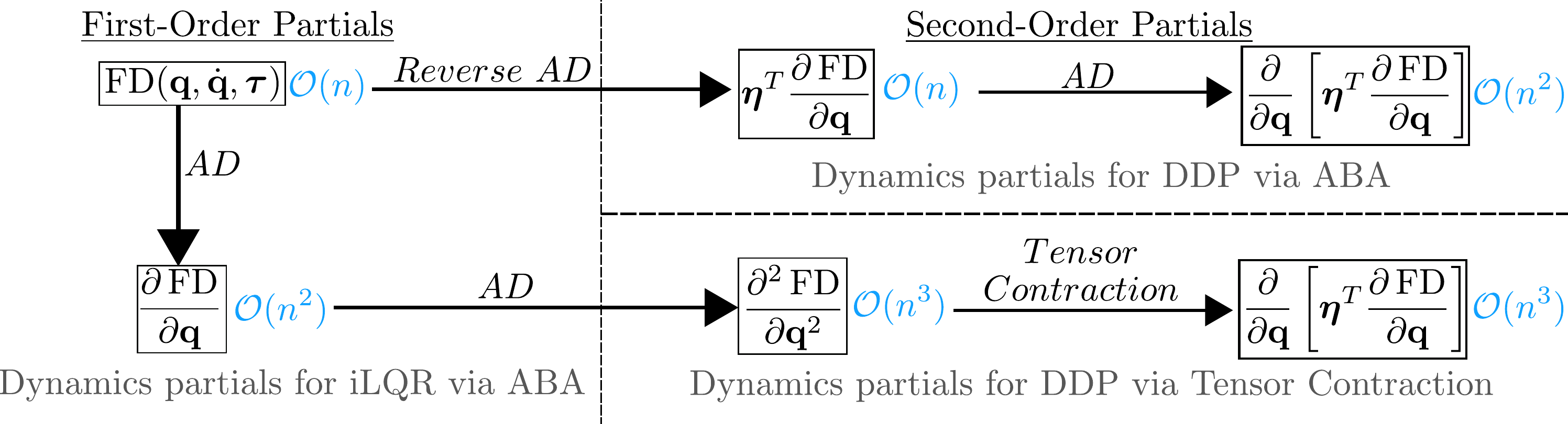}
    \caption{Computational approach for the partials needed in {\emph \ABAfirst}, {\emph \ABASecond}, and {\emph \TensorCont} methods}
    \vspace{-2.5ex}
    \label{fig:CompApproachABA}
\end{figure}

\quotelabel{rev: ComplexityClarify}{\rev{To motivate our approach for accelerating the evaluation of \eqref{eqn: Hqq}, consider any vector-valued function $\g(\x):\R^n \rightarrow \R^m$. 
Given any fixed vector $\ggamma \in \R^m$, reverse-mode derivative accumulation \cite{grievank2000principles} provides an approach to evaluate the vector-Jacobian product
\begin{equation}
\ggamma^T \pd[\g]{ \x} = \sum_{j=1}^m \gamma_j \pd[g_j]{\x}  = \pd{\x}\left[\ggamma^T \g\right]
\label{eq:BackDiff}
\end{equation}
with the same complexity as evaluating $\g(\x)$. It is 
implemented in many AD packages (e.g., CasADi \cite{andersson2019casadi}). 
In practice, the cost of evaluating \eqref{eq:BackDiff} can be bounded above by a small constant times the cost of evaluating $\g(\x)$, and the  constant is often three to four~\cite[Section 3.3]{grievank2000principles}. 
}}


Returning to \eqref{eqn: Hqq}, since the partials of $\FwdDyn$ are contracted with the vector $\eeta$ on the left, the desired partials 
can be computed efficiently using reverse-mode approaches, as diagrammed in Fig.~\ref{fig:CompApproachABA}. Since $\FwdDyn$ can be calculated in \Ord{n}, reverse-mode AD can be used to compute $\eeta^T \pd[\FwdDyn]{ \q}$ in \Ord{n} operations as well. This result is then differentiated further, achieving the necessary result $ \pd{ \q} \left[\eeta^T \pd[\FwdDyn]{  \q} \right]$ in \Ord{n^2} operations -- the same complexity as the first-order partials for $\FwdDyn$ itself. When partials for DDP are obtained with this approach, we denote the method as {\emph \ABASecond} since ABA is first used to evaluate $\FwdDyn$. \quotelabel{rev:adhessian}{\rev{This general strategy also applies to any dynamic system $\x_{k+1} = \f(\x_k, \u_k)$ for computing $\pd[]{\x}[\llambda\T \fx] = \llambda \cdot \f_{\x\x}$ in DDP. While the use of reverse-mode accumulation to compute Hessians is a standard option in AD packages, its use to accelerate DDP here is new.}}




\subsection{\rev{First-Order Derivatives of Rigid-Body Dynamics}}\label{sect:FirstOrderPartials}
The structure of rigid-body dynamics further enables efficiency gains for derivative evaluation. 
The iLQR and DDP algorithms require first-order derivatives of the dynamics, and these can be computed in  $\Ord{n^2}$ operations with AD tools applied to the ABA. 
When dynamics derivatives are computed in this manner, the resulting iLQR algorithm is denoted as {\emph \ABAfirst}. \revNew{This approach is diagrammed on the left side of \figref{fig:CompApproachABA}}. \quotelabel{rev: sentenceStructure}{\rev{ Derivatives of $\FwdDyn$ are, however, analytically related to the derivatives of $\InvDyn$ by \cite{carpentier2018analytical,jain1993linearization}
\begin{align}\label{eqn:FirstIDtoFirstFD}
\left. \pd[\FwdDyn]{\z} \right|_{\q_0,\qd_0,\t_0} &= -\M^{-1}(\q_0) \left. \pd[\InvDyn]{\z} \right|_{\q_0,\qd_0,\t_0}\quad{\textrm {and}}\\[.5ex]
\left. \pd[\FwdDyn]{\ttau} \right|_{\q_0,\qd_0,\t_0} &= \M^{-1}(\q_0) = \left( \left. \pd[\InvDyn]{\qdd}\right|_{\q_0,\qd_0,\qdd_0} \right)^{-1}\,,
\end{align}
where $\z$ can either represent the vector $\q$ or $\qd$, and where ${\q_0,\qd_0,\qdd_0}$, and $\t_0$ refer to the point of linearization.}}

\quotelabel{rev: ilQR_RNEA}{\rev{ 
These relationships provide alternate methods  for the first-order partials of $\FwdDyn$, \revNew{shown on the left side of Fig.~\ref{fig:DerivsRNEAgraph}}. The term $\pd[\InvDyn]{\z}$ can be computed in \Ord{n^2} complexity with AD tools or specialized algorithms (e.g., \cite{carpentier2018analytical,lee2005newton,sohl2001recursive,jain1993linearization}). The explicit computation of the mass matrix inverse can be avoided in \eqref{eqn:FirstIDtoFirstFD} by instead applying it indirectly via $n$ calls to the ABA algorithm (\Ord{n}) with the columns of $\pd[\InvDyn]{\z}$ as inputs for $\t$. This approach evaluates \eqref{eqn:FirstIDtoFirstFD} with total complexity $\Ord{n^2}$. Since \eqref{eqn:FirstIDtoFirstFD} relies on RNEA to obtain $\InvDyn$, we denote this method as {\emph \RNEAfirst}. As an alternate approach, the partials of $\FwdDyn$ are computed via \eqref{eqn:FirstIDtoFirstFD} with $O(n^3)$ complexity as follows. The partials of $\InvDyn$ are computed with $O(n^2)$ complexity, the mass matrix inverse is computed once with $O(n^2)$ complexity (e.g., via \cite{carpentier2018Inverse}), and a dense matrix-matrix multiply in \eqref{eqn:FirstIDtoFirstFD} finally sets the complexity at $O(n^3)$. Since matrix multiplications are optimized on modern hardware, this approach \cite{carpentier2018analytical} can be faster than the lower-order one.} We next extend the relation \eqref{eqn:FirstIDtoFirstFD} to the second-order case. }
\subsection{\rev{Second-Order Partials of Rigid-Body Dynamics}}\label{sect:SecondOrderPartials}

\begin{figure}[t]
    \centering
    \includegraphics[width = \columnwidth]{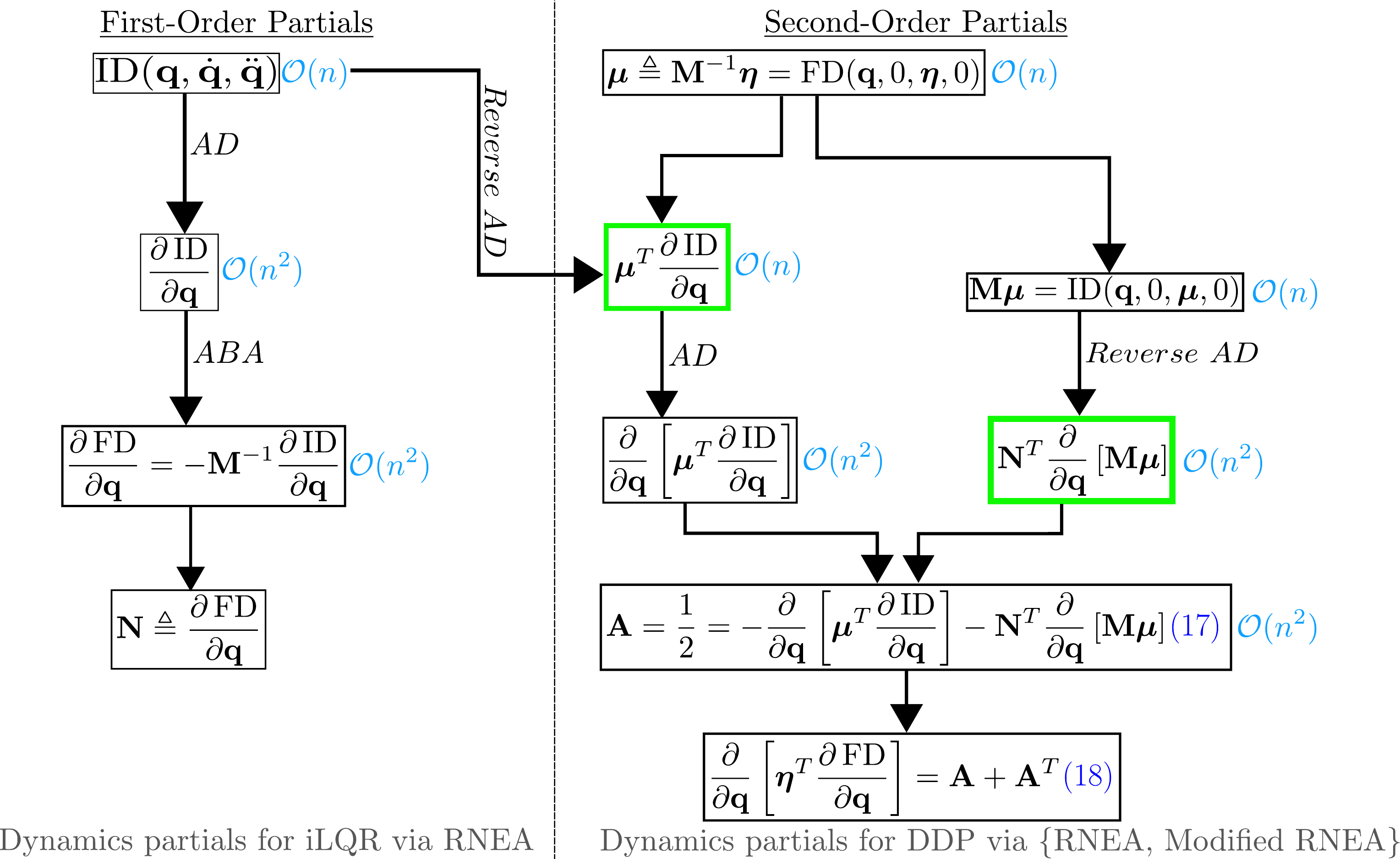}
    \vspace{-1ex}
    \caption{Computational approach for the partials needed in {\emph \RNEAfirst}, {\emph \RNEASecond}, and {\emph \ModRNEASecond}. For {\emph \ModRNEASecond}, the highlighted blocks~(in green) can be accelerated using the Modified RNEA~(see \algoref{algo:modRNEA})}
    \vspace{-5.5ex}
    \label{fig:DerivsRNEAgraph} 
\end{figure}

As the main technical contribution of the paper, this section presents an efficient way to include the second-order dynamics partials in DDP by employing reverse-mode AD and the relationship between first-order sensitivities~\eqref{eqn:FirstIDtoFirstFD}. 
\revNew{Figure \ref{fig:DerivsRNEAgraph} is diagrammed as a companion road-map to the following technical development.}
The derivation makes use of the identity 
\begin{equation}\label{eqn:ID2derivs}
\pdd[\InvDyn]{q_i}{\qdd} = \pd{ q_i} \M(\q) = \pd[\M]{ q_i}
\end{equation}
and the fact that for a fixed $\q$, $\qd$, and $\t$,  $\InvDyn(\q,\qd,\qdd)$ is implicitly dependent on $\FwdDyn$ through composition via $\qdd=\FwdDyn(\q,\qd,\t)$. Therefore, the  partials of $\InvDyn$ include the partials of $\FwdDyn$ through chain rule. 
We start by using reconsidering the second order partial in \eqref{eqn: Hqq} via the use of \eqref{eqn:FirstIDtoFirstFD}:
\begin{align}\label{eqn:aba_qq_a}
&\eeta^T \pd{q_j} \left[ \pd[\FwdDyn]{q_i} \right]= - \eeta^T \pd{q_j} \left[\M^{-1} \left.\pd[\InvDyn]{q_i} \right|_{\q,\qd,\qdd =\FwdDyn(\q,\qd,\t)} \right] \nonumber \\[.5ex]
&~~~= \eeta^T  \M^{-1} \left[ \pd[\M]{ q_j} \M^{-1} \pd[\InvDyn]{q_i}  - \pdd[\InvDyn]{q_i}{q_j} - \pdd[\InvDyn]{q_i}{\qdd} \pd[\FwdDyn]{q_j} \right]
\end{align}
\quotelabel{rev:negsign}{\rev{where we used that $\partial \M^{-1}/ \partial q_j = - \M^{-1} ( \partial \M / \partial q_j ) \M^{-1}$}}, and the last term in the overall result is from the chain rule for the second-order partials of $\InvDyn$. 
The identity~\eqref{eqn:ID2derivs} allows for the rewrite of~\eqref{eqn:aba_qq_a} such that 
\begin{align}
&\eeta^T \pd{q_j} \left[ \pd[\FwdDyn]{q_j} \right] \label{eq:SOP} \\ &~~~=\eeta^T  \M^{-1} \left[ \pd[\M]{q_j} \M^{-1} \pd[\InvDyn]{q_i} -  \right. \left. \pdd[\InvDyn]{q_i}{q_j} - \pd[\M]{q_i} \pd[\FwdDyn]{q_j} \right].  \nonumber
\end{align}
The matrix-vector product  $\M^{-1} \eeta$ can be computed efficiently using the ABA algorithm by ignoring gravity and the Coriolis term and using $\eeta$ as an input in place of $\t$. \quotelabel{rev:mufix_part1}{\rev{Once $\q$ and $\eeta$ are given, we fix the value of $\mmu \triangleq \M^{-1} \eeta = \FwdDyn(\q,0,\eeta, 0)$ (i.e., once fixed it no longer depends on $\q$). }}

We also fix each $\nnu_i \triangleq \frac{\partial \FwdDyn}{\partial q_i}$, and re-use this information in \eqref{eq:SOP} as follows
\begin{equation*}
\begin{aligned}
&\eeta^T \pd{q_j} \left[ \pd[\FwdDyn]{q_j} \right]\\ 
&~~~= -\mmu^T \pd{q_j} \left[ \M \nnu_i \right] - \pd{q_j} \left[\mmu^T \pd[\InvDyn]{q_i} \right]   -  \mmu^T \pd{ q_i} \left[ \M \nnu_j \right].
\end{aligned}
\end{equation*}
\quotelabel{rev:mufix}{\rev{Note that the key reorganization $\pd{q_j} [\mmu^T \pd[\InvDyn]{q_i}] = \mmu^T\pdd[\InvDyn]{q_i}{q_j}$ is possible since $\mmu$ is treated as constant after its initial evaluation.}} The resulting equation has a unique structure in that the terms $ \pd{ q_i} \left[ \M \nnu_j \right]$ and $\pd{q_j} \left[ \M \nnu_i \right] $ are related by index permutation. This observation allows us to define  $\mathbf{N} \triangleq \pd[\FwdDyn]{\q}$ such that 
\[
\mmu^T \pd{q_j} \left[ \M \nnu_i \right] =\left[ \mathbf{N}^T \pd{\q} \left[ \M \mmu \right] \right]_{ij} \;.
\]
As a result, the following definition 
\begin{equation}
 \mathbf{A} := -\frac{1}{2} \pd{\q} \left[ \mmu^T \pd[\InvDyn]{\q} \right] - \mathbf{N}^T \pd{\q} \left[ \M \mmu \right] 
 \label{eq:A_definition}
\end{equation}
provides 
\begin{equation}\label{eqn:FD_qq}
\pd{ \q} \left[ \eeta^T \frac{\partial \FwdDyn}{\partial \q} \right] = \A + \A^T.
\end{equation}

Both terms in $\A$ can be computed in \Ord{n^2} as follows. For the $\pd[]{\q} \left[ \mmu^T \pd[\InvDyn]{\q} \right]$ term, $\mmu^T \pd[\InvDyn]{\q} $ can first be computed using reverse-mode AD, then AD can be used again for the derivative of that result (Fig.~\ref{fig:DerivsRNEAgraph}). For the $ \mathbf{N}^T \pd{\q} \left[ \M \mmu \right]$ term, RNEA can be used to evaluate $\M \mmu = \InvDyn(\q,0,\mmu,0)$ by ignoring gravity and the Coriolis terms. 
This calculation is an \Ord{n} operation.
Considering the $n$ columns of $\N$ to contract on the left of $\N^T \pd{\q} \left[ \M \mmu \right]$ provides an \Ord{n^2} method via reverse-mode AD. The first-order partials in $\N$ have to be computed for any iLQR/DDP method, and are thus not an additional cost. 
%
Even though the formulas \eqref{eq:A_definition}, \eqref{eqn:FD_qq} result in the same complexity as applying reverse-mode AD to ABA, the new approach is faster since RNEA is simpler than ABA. 

Formulas for the other second-order partials are as follows.
Using the same approach as before, we can show that
\[
\pd{\qd} \left[ \eeta ^T \pd[\FwdDyn]{\qd } \right] = - \pd{\qd} \left[ \mmu^T \pd[\InvDyn]{\qd} \right] \,.
\]
For the mixed partials of $\q$ and $\qd$, we can show that
\[
\pd{\q}\left[ \eeta^T  \pd[\FwdDyn]{\qd} \right] = - \Ppsi\T \pd{\q} \left[ \M(\q) \mmu \right]  - \pd{\q} \left[ \mmu\T  \pd[\InvDyn]{\qd} \right]\,,
\]
where $\Ppsi = \pd[\FwdDyn]{\qd}$. Finally, for the mixed partials of $\q$ and $\t$, the resulting relationship is
\[
{\pd{\q}\left[ \eeta\T \pd[\FwdDyn]{\ttau} \right] = -\Xxi\T \pd{\q} \left[ \M(\q)  \mmu \right] }   \,,
\]
where $\Xxi = \pd[\FwdDyn]{\ttau}$. 
Using reverse-mode tools, all these partials can be computed with \Ord{n^2} complexity. We denote DDP algorithms that use this approach as {\emph \RNEASecond}.

Across these derivations, the term
$\mmu^T \pd[\InvDyn]{\q}$ is common.  
To evaluate $\mmu^T \pd[\InvDyn]{\q}$ we can either (1) use reverse AD  with the RNEA to calculate result or (2) provide a method to compute the term $\mmu^T \InvDyn$, and then calculate its gradient with AD. While the two approaches are similar, we present a refactoring of the RNEA to reduce computation requirements for $\mmu^T \InvDyn$ before applying AD. 


\subsection{Modified RNEA Algorithm}
\label{sec:modRNEA}
Here, we follow spatial vector algebra, notation, and body numbering conventions in~\cite{featherstone2014rigid}. Consider a poly-articulated tree-structured system of $N_B$ rigid bodies.  We denote $p(i)$ as the parent body of body $i$, and use the notation $j \succeq i$ to indicate when body $j$ is after body $i$ in the kinematic tree. Sums over pairs of related bodies can be carried out in either of the following ways:
\begin{equation}
 \sum_{i=1}^{N_B} \, \sum_{j |j \succeq i}  = \sum_{j=1}^{N_B} \, \sum_{i | i \preceq j } \, .
 \label{eq:sumIdentity}
\end{equation}

The modified RNEA output $
\mmu^T \InvDyn$ satisfies $ \mmu^T \t = \sum_{i=1}^N \mmu_i^T \t_i$. The torque $\t_i$ at joint $i$ is given as
\[
\t_i = \Pphi_i^T \sum_{j|j \succeq i}  \XM{j}{i}^T \mathbfcal{F}_j\,,
\]
where $\Pphi_i$ gives joint $i$'s free-modes, $\mathbfcal{F}_j =\I_j \a_j + \crf{\v_j} \I_j \v_j$ the inertial force of body $j$, $\v_j$ its spatial velocity, $\a_j$ its spatial acceleration, and $\times^*$ a cross product for spatial vectors \cite{featherstone2014rigid}. The  term $\mmu^T \t$ then satisfies
\[
\mmu^T \t = \sum_{i=1}^{N_B}  \sum_{j|j \succeq i} \mmu_i^T \Pphi_i^T \XM{j}{i}^T \mathbfcal{F}_j.
\]
The summation above is then refactored using \eqref{eq:sumIdentity} as 
\begin{equation}
    \mmu^T \InvDyn = \sum_{j=1}^{N_B} \underbrace{ \Big[ \sum_{i|i \preceq j} \XM{j}{i} \Pphi_i \mmu_i   \Big]^T }_{:=\w_j}   \mathbfcal{F}_j.
\end{equation}
This refactoring leads to the modified RNEA algorithm~(see \algoref{algo:modRNEA}). The result is that $\mmu\T \InvDyn$ can be computed in a single pass rather than two passes as in RNEA. The algorithm is still an \Ord{n} algorithm but leads to a simpler computation graph for reverse-mode AD. Whenever this method is used in DDP, it is referred to as {\emph \ModRNEASecond}. The computation workflow in this case is given as in Fig.~\ref{fig:DerivsRNEAgraph}, where the modified RNEA is used to accelerate the blocks highlighted in green.

\input{modRNEAalgo.tex}

\subsection{Summary}\label{sect: summary}
Before proceeding to the presentation of comparative results, we briefly review the methods introduced for incorporating partials into iLQR and DDP. When only using  first-order dynamics partials, we have {\emph \ABAfirst} and {\emph \RNEAfirst}. For full second-order methods, we have {\emph \ABASecond}, {\emph \RNEASecond}, and {\emph \ModRNEASecond} methods, all avoiding explicitly computing the second-order dynamics derivative tensor. Finally the conventional DDP method {\emph \TensorCont} involves computing the second-order derivative tensor. The computation approaches of these methods have been diagrammed in \figref{fig:CompApproachABA}
and \figref{fig:DerivsRNEAgraph}.

\section{Results}

\begin{figure}[t]
    \centering
    \includegraphics[width=.9\columnwidth]{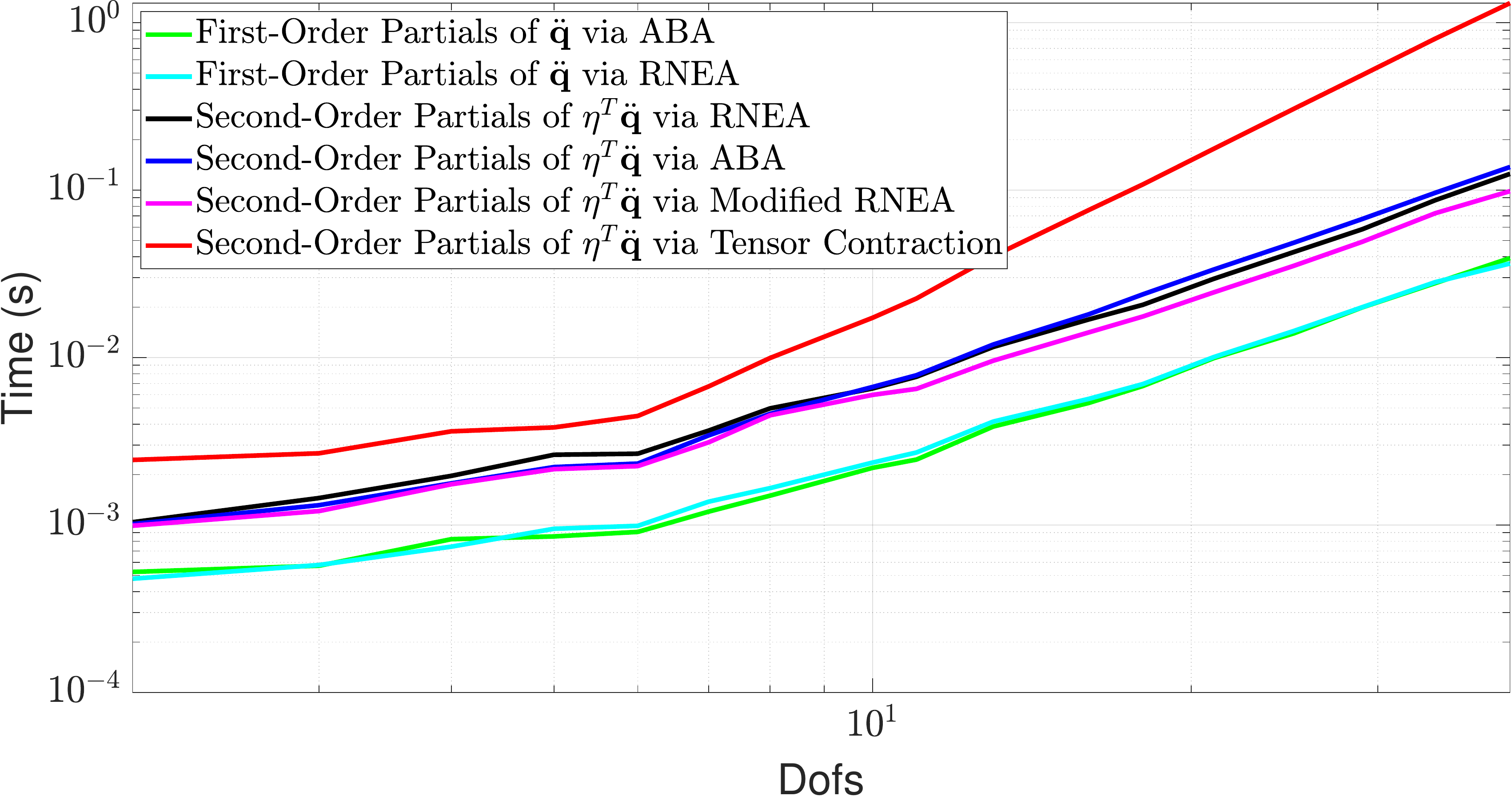}
    \caption{\rev{The computation time for evaluating the partials of the dynamics in {\sc Matlab} with {CasADi}.}}
    \label{fig:Derivs}
\end{figure}

To test the performance and scalability of our proposed methods, we evaluate their application to solve an OCP for an \rev{underactuated}  $n$-link pendubot. The trajectory optimization problem here was a swing-up problem and this goal was encoded via design of running and terminal costs. These costs were similar for all proposed methods regardless of the number of links in the model. As the number of links is increased, their nonlinear couplings on one another present additional challenge for solving the OCP. To further make the control problem challenging, the final link in the system was left un-actuated. 
For all the proposed methods, the same convergence criteria was used with convergence indicated by a negligible ($<10^{-9}$) reduction in the  cost function between iterations. 
We first compare the computation time of the dynamics partials~(\secref{sect:Results_derivs}) using the methods described previously and then compare the addition of those partials within DDP/iLQR optimization frameworks as appropriate~(see \secref{sect:Results_control}). This work was implemented in {\sc Matlab}\footnote{\revNew{Open-source code:  \url{https://tinyurl.com/468ynkuu}}} alongside the { CasADi} \cite{andersson2019casadi} Toolkit which allows for rapid and efficient testing of AD approaches. Since the partials are evaluated in the {CasADi} virtual machine through {\sc MATLAB}, merits of the methods should be assessed via comparison between them, while future work will study improving absolute timing numbers via C/C++ implementation.

\subsection{Dynamics Partials}\label{sect:Results_derivs}
Within DDP/iLQR, the computation of the partials of the dynamics is the most computationally expensive part of the optimization process. Figure \ref{fig:Derivs} compares the computation time of the different methods for evaluating these partials. As shown, evaluation of the second-order partials by tensor contraction takes the longest time whereas all other second-order partials have the same computational complexity as the first-order dynamic partials (as indicated by the slope on the log-log plot). The most competitive second-order approach 
requires approximately only $2.5$ times more computation time than first-order partials. Second-order partials via RNEA/modified RNEA were faster than second-order partials via ABA since RNEA is simpler than ABA, with the modified RNEA outperforming RNEA.
These results indicate that when compared to conventional second-order tensorial dynamic partials, the proposed methods have the potential to reduce the computational overhead of DDP to be competitive with iLQR. 
\begin{figure}[tb]
    \centering
    \includegraphics[width=.9\columnwidth]{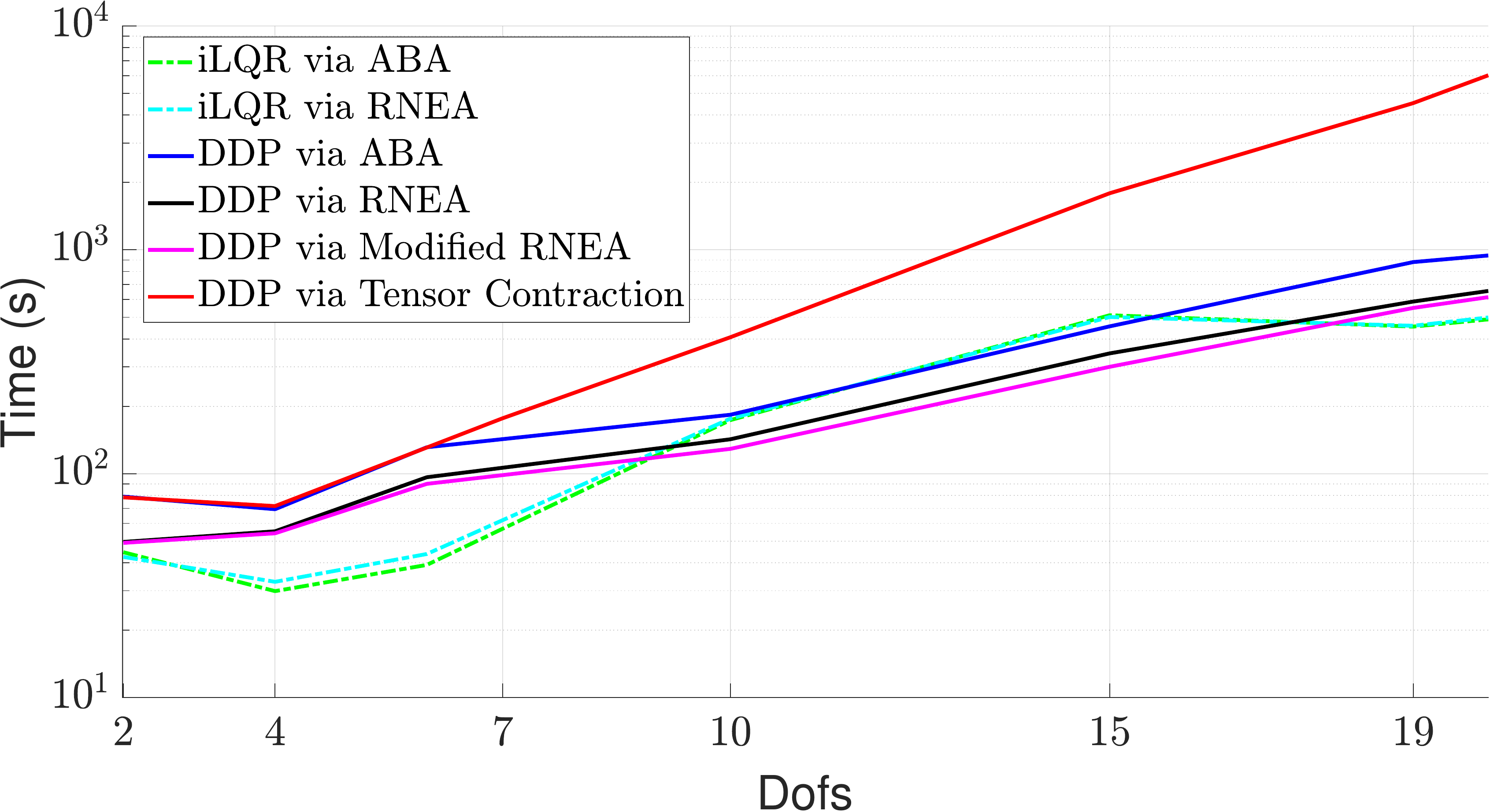}
    \caption{\rev{Computation time for solving a swing-up OCP for an $n-$link pendubot in {\sc Matlab}.}}
    \label{fig:LogTime}
    \vspace{-0.5ex}
\end{figure}

\subsection{Trajectory Optimization: DDP/iLQR Framework}\label{sect:Results_control}
We then include the dynamics partials in DDP/iLQR as appropriate and evaluate the OCP for the pendubot. Note that for an OCP with a horizon of $N$ timesteps, each iteration of iLQR/DDP must evaluate the derivatives $N$ times, motivating the need for their rapid evaluation.  Figure \ref{fig:LogTime} illustrates the time required to solve the OCP to convergence with DDP and iLQR variants. Tensor-free DDP variants had evaluation times comparable to the iLQR ones, while {\em \TensorCont} takes longer to optimize due to its higher cost of evaluating derivatives. 
\revNew{The comparative performance of the fastest iLQR and DDP variants depends on the problem instance. Trajectory updates performed for either algorithm depend on the non-linearity of the system considered, and on the initial guess, which prevents uniformly recommending iLQR or DDP over the other.}

\begin{figure}[bt]
    \centering
    \includegraphics[width=.8\columnwidth]{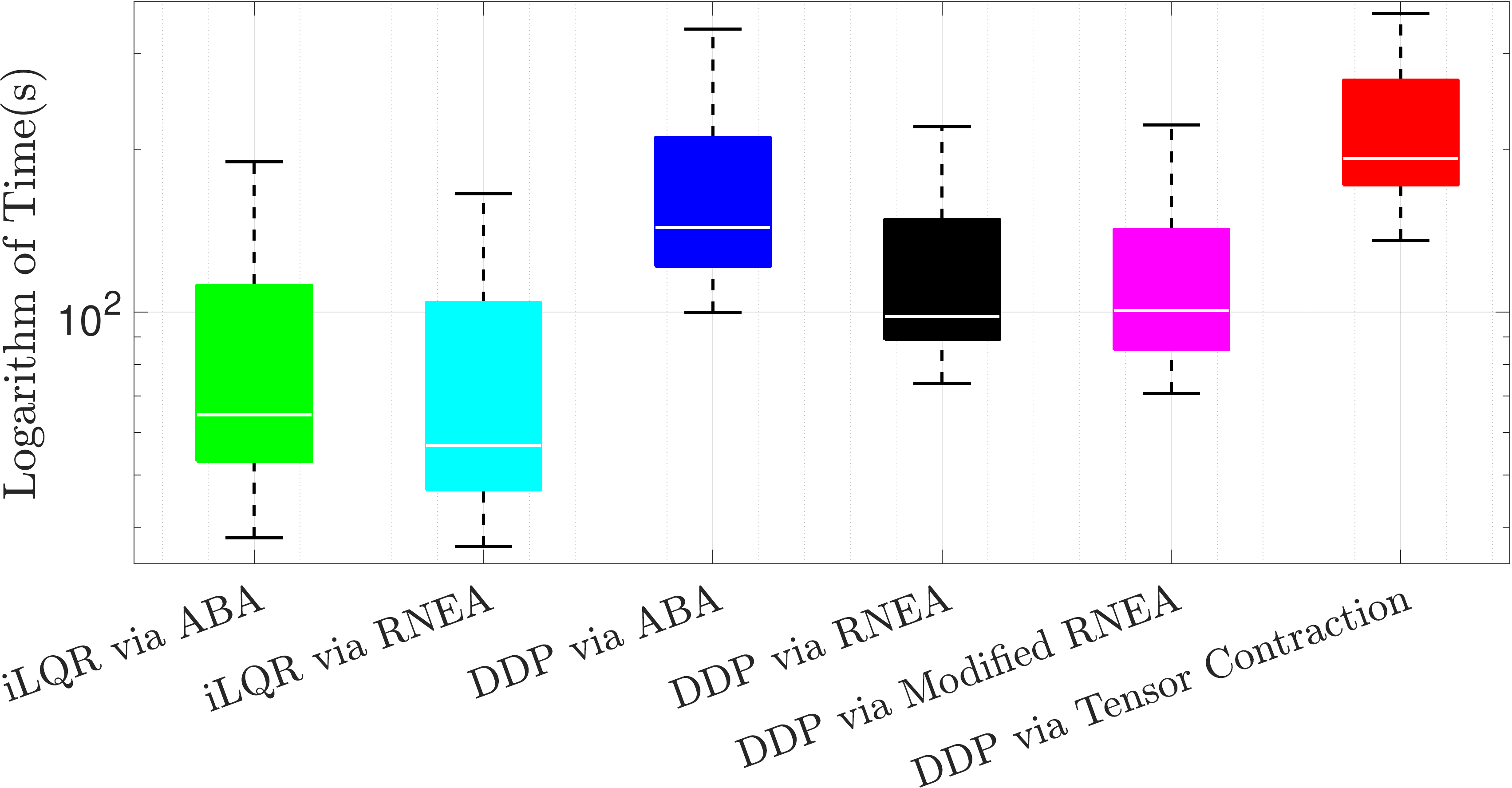}
    \vspace{-2ex}
    \caption{\rev{Time to converge for a $7$-link model with a random sampling of the initial conditions around the downward configuration of the pendubot.}}
    \label{fig:boxplotTime}
\end{figure}

\rev{We also evaluate the mean time to compute the DDP/iLQR variants for a $7$-link pendubot model. We use a dissipative controller as the initial control trajectory and randomize the initial state vector around the downward configuration of the pendubot.} This is a difficult problem, as it forces the control to pump energy into the system in order to drive it to the upright configuration. Figure \ref{fig:boxplotTime} illustrates the time to solve the OCP with DDP/iLQR variants. As shown, 
{\emph \TensorCont} took the longest time to converge, whereas the tensor-free DDP strategies took more time compared to iLQR in this case. A portion of the additional time for DDP is attributed to repeats of the backward sweep due to the regularization needed for DDP in this case. While this motivates focus on these aspects in future work, it is worth noting that the running cost in this case was convex, and this provides benefit to iLQR in terms of avoiding regularization.

\begin{figure}[t]
    \centering
    \includegraphics[width=.8\columnwidth]{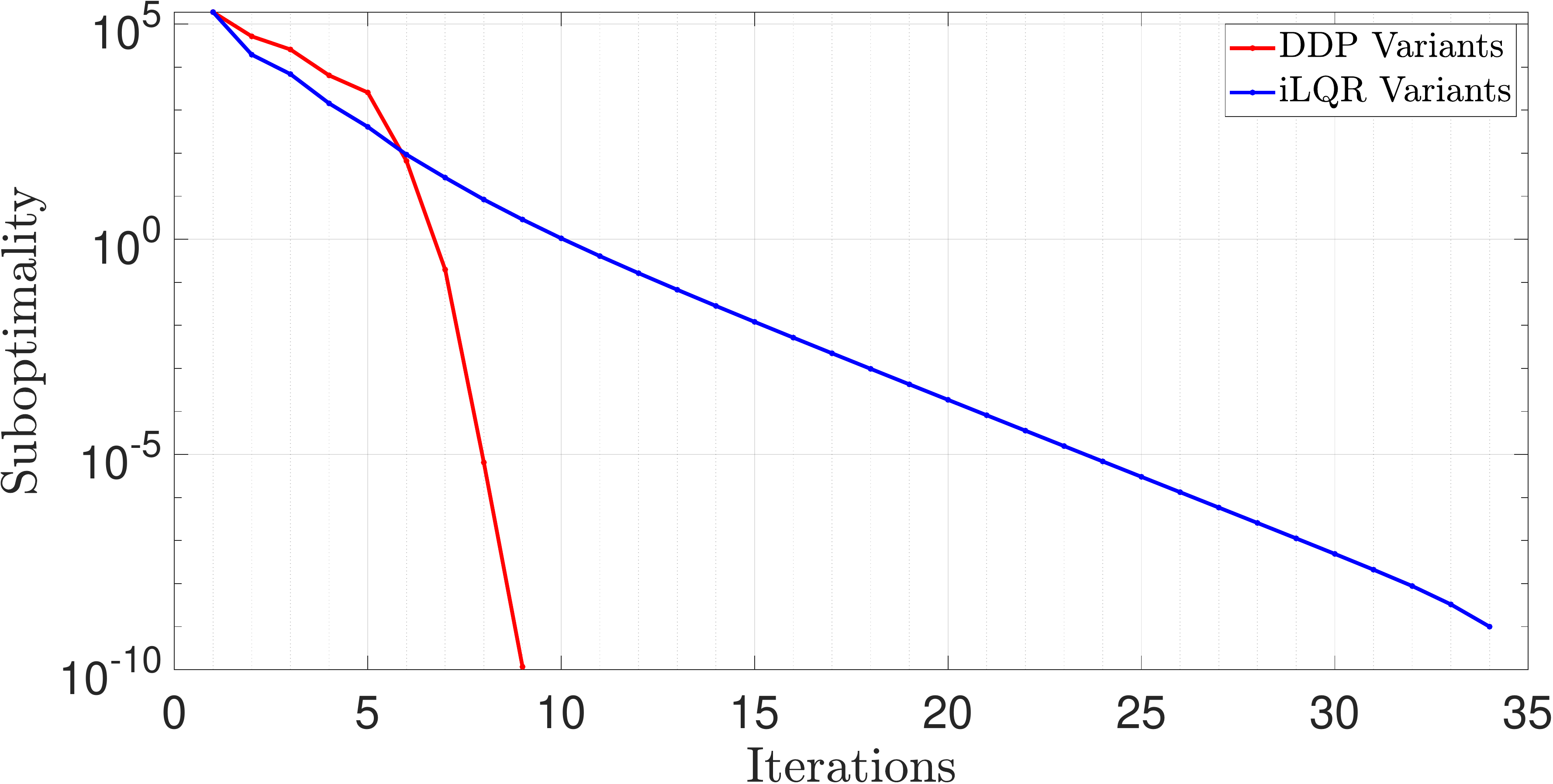} 
    \caption{Cost convergence over iterations for a $7-$link KUKA LBR model.}
    \label{fig:SuboptimalityIteration}
    \vspace{-1.5ex}
\end{figure}

Finally, we optimize a trajectory for a $7-$link KUKA LBR manipulator (available in {\sc Matlab}'s Robotics Toolbox) using DDP/iLQR methods.  Figure~\ref{fig:SuboptimalityIteration} illustrates suboptimality vs. iterations for iLQR and DDP when applied to the manipulator. The suboptimality measures the difference between the current cost function value and its value at the end of the iterations. As illustrated, the DDP variants featured quadratic convergence whereas the iLQR variants featured super-linear convergence. 
This figure also illustrates that the DDP variants took fewer iterations compared to iLQR counterparts. 

\begin{figure}[t]
    \centering
    \includegraphics[width = \columnwidth]{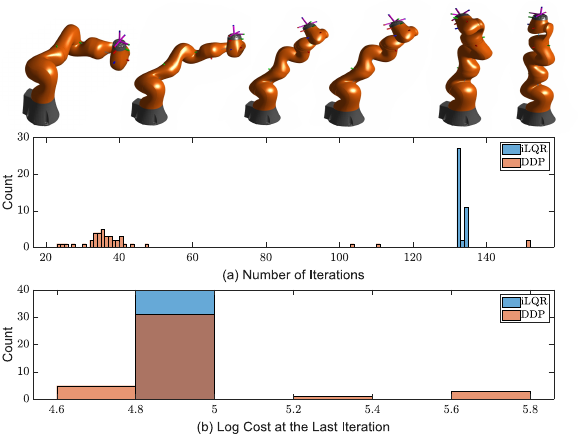}
    \vspace{-2ex}
    \caption{\rev{Optimization results for a $7-$link KUKA LBR  model with randomized initial control inputs from an Ornstein-Uhlenbeck noise process. The LBR was simulated using {\sc Matlab}'s Robotics toolbox.
    }}
    \label{fig:RandomCtrl}
    \vspace{-1.5ex}
\end{figure}
\quotelabel{rev: ROA_Q}{\rev{
Since the solution of an OCP is dependent on the initial conditions, we randomize the initial controller using Ornstein-Uhlenbeck process noise for the $7-$link  KUKA LBR manipulator and solved the OCP problem using either iLQR/DDP in $40$ separate instances. The first pane of \figref{fig:RandomCtrl} illustrates the progression of that manipulator from an initial configuration to the balanced upright configuration along an optimal trajectory. Subplot (a) of \figref{fig:RandomCtrl} illustrates the empirical probability density function (pdf)  of the number of iterations over those optimizations. Subplot (b) illustrates the pdf of the log of the final cost of each optimal solution. As illustrated in subplot (a), in $95\%$ of the simulations, iLQR had a higher number of iterations. On average, iLQR had three times as many iterations as DDP.
From subplot (b), we note that most of the simulations regardless of DDP/iLQR converged to similar solution; in fact, DDP converged to a different solution than iLQR in only three instances. Figure \ref{fig:RandomCtrl} illustrates that the inclusion of second-order information in DDP will result in similar converged solution and in fewer iterations. 
This result is powerful in that  
the addition of second-order information results in algorithm that converges in less iterations and to the same minima.
}}

\subsection{Discussion}
As compared to conventional approaches to DDP that explicitly compute second-order derivatives of the dynamics, the proposed DDP variants presented herein show a marked improvement in their computational complexity and computation time. This is especially important as second-order information retains better local fidelity to the original model and therefore a second-order approximation better captures the nonlinear effects of the system. This feature is expected to be important for complex systems such as quadrupeds whose coupled nonlinear dynamics might not be accurately captured by a first-order approximation. This second-order information could be of value in critical circumstances, for example, the increased fidelity could help a quadruped prevent falls or handle disturbances when traversing unstructured terrain. 

We noted previously that DDP includes a regularization scheme to ensure that $\Quu$ remains positive-definite during the backward pass. This regularization incurs additional computational cost as compared to iLQR methods, and this additional cost cannot be anticipated before running the optimization. In~\figref{fig:boxplotTime}, the proposed DDP variants needed more computation time than the iLQR variants, though we expected those evaluation times to be more similar based on other tests (Fig.~\ref{fig:Fig1}). We attribute a portion of the computational cost to repeats of the backward sweep when regularization fails, as most of the computation in DDP is spent evaluating the backward sweep. 

Lastly, our results showed that iLQR typically has more iterations but overall had comparable computation time as proposed DDP variants. Therefore, the benefits of DDP can be included in trajectory optimization without the previously significant sacrifice in evaluation time, and can be done in fewer iterations. This may lead to more robust model-predictive  control wherein warm staring online may keep DDP iterations within the quadratic convergence well. 

\section{Conclusion and Future Directions}\label{sect: Conclusions}
\quotelabel{rev: ConclusionAndFuture}{\rev{
This work has made use of reverse-mode AD tools to quickly evaluate different approaches for obtaining the derivatives needed by DDP. We also extended the relationship between the first-order sensitivities of $\InvDyn$ and $\FwdDyn$ \cite{carpentier2018analytical,jain1993linearization} for rigid-body systems to the second-order case. The combination of this new approach and AD tools allows for the evaluation of the needed derivatives in DDP with the same complexity as iLQR. Lastly, we introduce a restructuring of RNEA to derive a modified RNEA that returns $\mu\T \InvDyn$ in $\Ord{n}$ complexity, and enables the fastest DDP algorithm.}}

\quotelabel{rev:otherFuture}{\rev{
While AD tools are convenient, they are general purpose, and thus may not be optimal. Alternative analytical methods for taking derivatives of rigid-body dynamics can accumulate the derivatives recursively, as in~\cite{carpentier2018analytical,lee2005newton,sohl2001recursive,jain1993linearization}. Recently, we extended the modified RNEA algorithm with an analytical accumulation of its first-order partials in a reverse-mode fashion, and further evaluation of this result is of immediate interest. Moreover, we aim to extend this work to address rigid-body dynamics with contacts \cite{DDPwithContacts,li2020hybrid} by using similar approaches as in this paper. This generalization would then allow for use with hybrid dynamic systems that arise in legged locomotion problems.

There are many other opportunities that this work motivates as next steps. While we noted that the presented work is general for any explicit integration scheme, we also see opportunity to extend this work for implicit integration and implicit DDP~\cite{ImplicitDDP}. 
Further, while the DDP used here was a single-shooting solver, our contributions  could be used in multi-shooting DDP and other numerical optimal control solvers. }}
\quotelabel{rev:softRobotFuture}{\rev{Finally, our work may find applicability when working to control soft robots. Many multi-segment soft-body robots can be modeled assuming piece-wise constant curvature (PCC)~\cite{softToRigid}, approximating PCC with a high-DoF augmented rigid model \cite{softToRigid}, or considering discrete Cosserat models~\cite{DiscreteCosseratSoftRobots}. The ideas herein may find applicability for these models due to their dynamics equations taking a similar structural form (e.g., \eqref{eq:id}) as rigid-body systems. Augmented rigid models are most directly applicable in 2D, but  generalizations of the RNEA and ABA \cite{DiscreteCosseratSoftRobots} for the 3D continuum case also present interesting future avenues to broaden the application of this work. 
}}









\bibliographystyle{IEEEtran}
\bibliography{Resources/Cited.bib}
\end{document}

%% file: modRNEAalgo.tex
\begin{algorithm}[b]
\caption{Modified RNEA Algorithm}
\label{algo:modRNEA}
\begin{algorithmic}[1]
\REQUIRE $\q, \qd,\qdd, \mmu, model$
\STATE $\v_0 = 0, \a_0 = -\a_g ,\w_0=0, s=0$
\FOR{$i=1$ to $N$}
\STATE $\v_i = \XM{i}{p(i)}\,\v_{p(i)}+\Pphi_i \,\qd_i$\\[.5ex]
\STATE $\w_i = \XM{i}{p(i)}\,\w_{p(i)}+\Pphi_i \,\mmu_i$\\[.5ex]
\STATE $\a_i = \XM{i}{p(i)}\, \a_{p(i)} + \crm{\v_i}  \Pphi_i \qd_i + \Pphi_i \qdd_i$\\[.5ex]
\STATE $s += \w_i\T ( \I_i \a_i + \crf{\v_i} \I_i \v_i )$\\[.5ex]
\ENDFOR
\RETURN $s=\mmu^T \t$
\end{algorithmic}
\label{alg:cor}
\end{algorithm}

%% file: root.bbl
\begin{thebibliography}{10}
\providecommand{\url}[1]{#1}
\csname url@samestyle\endcsname
\providecommand{\newblock}{\relax}
\providecommand{\bibinfo}[2]{#2}
\providecommand{\BIBentrySTDinterwordspacing}{\spaceskip=0pt\relax}
\providecommand{\BIBentryALTinterwordstretchfactor}{4}
\providecommand{\BIBentryALTinterwordspacing}{\spaceskip=\fontdimen2\font plus
\BIBentryALTinterwordstretchfactor\fontdimen3\font minus
  \fontdimen4\font\relax}
\providecommand{\BIBforeignlanguage}[2]{{%
\expandafter\ifx\csname l@#1\endcsname\relax
\typeout{** WARNING: IEEEtran.bst: No hyphenation pattern has been}%
\typeout{** loaded for the language `#1'. Using the pattern for}%
\typeout{** the default language instead.}%
\else
\language=\csname l@#1\endcsname
\fi
#2}}
\providecommand{\BIBdecl}{\relax}
\BIBdecl

\bibitem{wieber2016modeling}
P.-B. Wieber, R.~Tedrake, and S.~Kuindersma, ``Modeling and control of legged
  robots,'' in \emph{Springer Handbook of Robotics}, 2016, pp. 1203--1234.

\bibitem{diehl2009efficient}
M.~Diehl, H.~J. Ferreau, and N.~Haverbeke, ``Efficient numerical methods for
  nonlinear {MPC} and moving horizon estimation,'' in \emph{Nonlinear model
  predictive control}.\hskip 1em plus 0.5em minus 0.4em\relax Springer, 2009,
  pp. 391--417.

\bibitem{mayne2000constrained}
D.~Q. Mayne, J.~B. Rawlings, C.~V. Rao, and P.~O. Scokaert, ``Constrained model
  predictive control: Stability and optimality,'' \emph{Automatica}, vol.~36,
  no.~6, pp. 789--814, 2000.

\bibitem{MultiShooting}
M.~Diehl, H.~G. Bock, H.~Diedam, and P.-B. Wieber, ``Fast direct multiple
  shooting algorithms for optimal robot control,'' in \emph{Fast motions in
  biomechanics and robotics}.\hskip 1em plus 0.5em minus 0.4em\relax Springer,
  2006, pp. 65--93.

\bibitem{Collocation}
M.~Kelly, ``An introduction to trajectory optimization: How to do your own
  direct collocation,'' \emph{SIAM Review}, vol.~59, no.~4, pp. 849--904, 2017.

\bibitem{centroidalDyn}
H.~Dai, A.~Valenzuela, and R.~Tedrake, ``Whole-body motion planning with
  centroidal dynamics and full kinematics,'' in \emph{IEEE-RAS International
  Conference on Humanoid Robots}, 2014, pp. 295--302.

\bibitem{koenemann2015whole}
J.~Koenemann, A.~Del~Prete, Y.~Tassa, E.~Todorov, O.~Stasse, M.~Bennewitz, and
  N.~Mansard, ``Whole-body model-predictive control applied to the {HRP}-2
  humanoid,'' in \emph{IEEE/RSJ Int.~Conf.~on Intelligent Robots and Systems},
  2015, pp. 3346--3351.

\bibitem{neunert2018whole}
M.~Neunert, M.~St{\"a}uble, M.~Giftthaler, C.~D. Bellicoso, J.~Carius,
  C.~Gehring, M.~Hutter, and J.~Buchli, ``Whole-body nonlinear model predictive
  control through contacts for quadrupeds,'' \emph{IEEE Robotics and Automation
  Letters}, vol.~3, no.~3, pp. 1458--1465, 2018.

\bibitem{grandia2019feedback}
R.~Grandia, F.~Farshidian, R.~Ranftl, and M.~Hutter, ``Feedback {MPC} for
  torque-controlled legged robots,'' in \emph{IEEE/RSJ Int.~Conf.~on
  Intelligent Robots and Systems}, 2019, pp. 4730--4737.

\bibitem{tassa2012synthesis}
Y.~Tassa, T.~Erez, and E.~Todorov, ``Synthesis and stabilization of complex
  behaviors through online trajectory optimization,'' in \emph{IEEE/RSJ
  Int.~Conf.~on Intelligent Robots and Systems}, 2012, pp. 4906--4913.

\bibitem{mayne1966DDP}
D.~Mayne, ``A second-order gradient method for determining optimal trajectories
  of non-linear discrete-time systems,'' \emph{International Journal of
  Control}, vol.~3, no.~1, pp. 85--95, 1966.

\bibitem{lantoine2012hybrid}
G.~Lantoine and R.~P. Russell, ``A hybrid differential dynamic programming
  algorithm for constrained optimal control problems. {Part} 1: {Theory},''
  \emph{Journal of Optimization Theory and Applications}, vol. 154, no.~2, pp.
  382--417, 2012.

\bibitem{howell2019altro}
T.~A. Howell, B.~E. Jackson, and Z.~Manchester, ``{ALTRO}: A fast solver for
  constrained trajectory optimization,'' in \emph{IEEE/RSJ Int.~Conf.~on
  Intelligent Robots and Systems}, 2019, pp. 7674--7679.

\bibitem{InteriorPointDDP}
A.~Pavlov, I.~Shames, and C.~Manzie, ``Interior point differential dynamic
  programming,'' \emph{IEEE Transactions on Control Systems Technology}, 2021.

\bibitem{li2020hybrid}
H.~Li and P.~M. Wensing, ``Hybrid systems differential dynamic programming for
  whole-body motion planning of legged robots,'' \emph{IEEE Robotics and
  Automation Letters}, vol.~5, no.~4, pp. 5448--5455, 2020.

\bibitem{tassa2014control}
Y.~Tassa, N.~Mansard, and E.~Todorov, ``Control-limited differential dynamic
  programming,'' in \emph{IEEE Int.~Conf.~on Robotics and Automation}, 2014,
  pp. 1168--1175.

\bibitem{sola2020squash}
J.~Marti-Saumell, J.~Sola, C.~Mastalli, and A.~Santamaria-Navarro, ``Squash-box
  feasibility driven differential dynamic programming,'' in \emph{IEEE/RSJ
  Int.~Conf.~on Intelligent Robots and Systems}, 2020.

\bibitem{pellegrini2017multiple}
E.~Pellegrini and R.~P. Russell, ``A multiple-shooting differential dynamic
  programming algorithm,'' in \emph{AAS/AIAA Space Flight Mechanics Meeting},
  vol.~2, 2017.

\bibitem{giftthaler2018family}
M.~Giftthaler, M.~Neunert, M.~St{\"a}uble, J.~Buchli, and M.~Diehl, ``A family
  of iterative {Gauss-Newton} shooting methods for nonlinear optimal control,''
  in \emph{IEEE/RSJ Int.~Conf.~on Intelligent Robots and Systems}, 2018, pp.
  1--9.

\bibitem{plancher2018performance}
B.~Plancher and S.~Kuindersma, ``A performance analysis of parallel
  differential dynamic programming on a {GPU},'' in \emph{Workshop on the
  Algorithmic Foundations of Robotics}, 2018, pp. 656--672.

\bibitem{farshidian2017real}
F.~Farshidian, E.~Jelavic, A.~Satapathy, M.~Giftthaler, and J.~Buchli,
  ``Real-time motion planning of legged robots: A model predictive control
  approach,'' in \emph{IEEE-RAS Int.~Conf.~on Humanoid Robotics}, 2017, pp.
  577--584.

\bibitem{li2020model}
H.~Li, R.~J. Frei, and P.~M. Wensing, ``Model hierarchy predictive control of
  robotic systems,'' \emph{IEEE Robotics and Automation Letters}, vol.~6,
  no.~2, pp. 3373--3380, 2021.

\bibitem{li2004iterative}
W.~Li and E.~Todorov, ``Iterative linear quadratic regulator design for
  nonlinear biological movement systems.'' in \emph{ICINCO}, 2004, pp.
  222--229.

\bibitem{grievank2000principles}
A.~Grievank, ``Principles and techniques of algorithmic differentiation:
  Evaluating derivatives,'' \emph{SIAM, Philadelphia}, 2000.

\bibitem{carpentier2018analytical}
J.~Carpentier and N.~Mansard, ``Analytical derivatives of rigid body dynamics
  algorithms,'' in \emph{Robotics: Science and Systems}, 2018.

\bibitem{jain1993linearization}
A.~Jain and G.~Rodriguez, ``Linearization of manipulator dynamics using spatial
  operators,'' \emph{IEEE Transactions on Systems, Man, and Cybernetics},
  vol.~23, no.~1, pp. 239--248, 1993.

\bibitem{originalRNEA1979}
D.~E. Orin, R.~McGhee, M.~Vukobratovi{\'c}, and G.~Hartoch, ``Kinematic and
  kinetic analysis of open-chain linkages utilizing {Newton-Euler} methods,''
  \emph{Math Biosciences}, vol.~43, no. 1-2, pp. 107--130, 1979.

\bibitem{featherstone2014rigid}
R.~Featherstone, \emph{Rigid body dynamics algorithms}.\hskip 1em plus 0.5em
  minus 0.4em\relax Springer, 2014.

\bibitem{OriginalMassMatrix}
M.~W. Walker and D.~E. Orin, ``Efficient dynamic computer simulation of robotic
  mechanisms,'' 1982.

\bibitem{liao1991convergence}
L.-Z. Liao and C.~A. Shoemaker, ``Convergence in unconstrained discrete-time
  differential dynamic programming,'' \emph{IEEE Transactions on Automatic
  Control}, vol.~36, no.~6, pp. 692--706, 1991.

\bibitem{andersson2019casadi}
J.~A. Andersson, J.~Gillis, G.~Horn, J.~B. Rawlings, and M.~Diehl, ``{CasADi}:
  a software framework for nonlinear optimization and optimal control,''
  \emph{Mathematical Programming Computation}, vol.~11, no.~1, pp. 1--36, 2019.

\bibitem{lee2005newton}
S.-H. Lee, J.~Kim, F.~C. Park, M.~Kim, and J.~E. Bobrow, ``Newton-type
  algorithms for dynamics-based robot movement optimization,'' \emph{IEEE
  Transactions on robotics}, vol.~21, no.~4, pp. 657--667, 2005.

\bibitem{sohl2001recursive}
G.~A. Sohl and J.~E. Bobrow, ``A recursive multibody dynamics and sensitivity
  algorithm for branched kinematic chains,'' \emph{J. Dyn. Sys., Meas.,
  Control}, vol. 123, no.~3, pp. 391--399, 2001.

\bibitem{carpentier2018Inverse}
\BIBentryALTinterwordspacing
J.~Carpentier, ``{Analytical Inverse of the Joint Space Inertia Matrix},''
  2018. [Online]. Available: \url{https://hal.laas.fr/hal-01790934}
\BIBentrySTDinterwordspacing

\bibitem{DDPwithContacts}
R.~Budhiraja, J.~Carpentier, C.~Mastalli, and N.~Mansard, ``Differential
  dynamic programming for multi-phase rigid contact dynamics,'' in
  \emph{IEEE-RAS Int.~Conf.~on Humanoid Robots}, 2018, pp. 1--9.

\bibitem{ImplicitDDP}
I.~Chatzinikolaidis and Z.~Li, ``Trajectory optimization of contact-rich
  motions using implicit differential dynamic programming,'' \emph{IEEE
  Robotics and Automation Letters}, vol.~6, no.~2, pp. 2626--2633, 2021.

\bibitem{softToRigid}
R.~K. Katzschmann, C.~Della~Santina, Y.~Toshimitsu, A.~Bicchi, and D.~Rus,
  ``Dynamic motion control of multi-segment soft robots using piecewise
  constant curvature matched with an augmented rigid body model,'' in
  \emph{IEEE Int. Conf. on Soft Robotics}, 2019, pp. 454--461.

\bibitem{DiscreteCosseratSoftRobots}
F.~Renda, F.~Boyer, J.~Dias, and L.~Seneviratne, ``Discrete cosserat approach
  for multisection soft manipulator dynamics,'' \emph{IEEE Transactions on
  Robotics}, vol.~34, no.~6, pp. 1518--1533, 2018.

\end{thebibliography}
